\documentclass[11pt,letterpaper]{article}
\usepackage{emnlp2017}
\usepackage{times}
\usepackage{latexsym}

\usepackage{xcolor}
\usepackage{listings}
\colorlet{light-gray}{gray!20}
\usepackage{graphicx}
\usepackage{array}
\usepackage{mathtools}
\usepackage{color,soul}
\usepackage{enumitem}
\usepackage{caption}
\usepackage[position=top]{subfig}
\emnlpfinalcopy

\def\emnlppaperid{***}

\title{Transfer Learning for Named-Entity Recognition with Neural Networks}

\author{Ji Young Lee\thanks{\hspace{3mm}These authors contributed equally to this work.}\\
	    MIT\\
	    {\tt jjylee@mit.edu}
	   \And
	Franck Dernoncourt\footnotemark[1]\\
   	MIT\\
   {\tt francky@mit.edu}
   \And
	 Peter Szolovits \\
   	MIT\\
   {\tt psz@mit.edu}
   }

\date{}

\begin{document}

\maketitle

\begin{abstract}
Recent approaches based on artificial neural networks (ANNs) have shown promising results for named-entity recognition (NER). In order to achieve high performances, ANNs
need to be trained on a large labeled dataset. However, labels might be difficult to obtain for the dataset on which the user wants to perform NER:
label scarcity 
is particularly pronounced for patient note de-identification, which is an instance  of NER.
In this work, we analyze to what extent transfer learning may address this issue. In particular, we demonstrate that transferring an ANN model trained on a large labeled dataset to another dataset with a limited number of labels improves upon the state-of-the-art results on two different datasets for patient note de-identification.
\end{abstract}

\section{Introduction}

Electronic health records (EHRs) have been widely adopted in some countries such as the United States and represent gold mines of information for medical research. 
The majority of EHR data exist in unstructured form such as patient notes~\cite{murdoch2013inevitable}.
Applying natural language processing on patient notes can improve the phenotyping of patients~\cite{ananthakrishnan2013improving,pivovarov2015automated,halpern2016electronic}, which has many downstream applications such as the understanding of diseases~\cite{liao2015development}.

However, before patient notes can be shared with medical investigators, some types of information, referred to as protected health information (PHI), must be removed in order to preserve patient confidentiality. In the United States, the Health Insurance Portability and Accountability Act (HIPAA)~\cite{office2002standards} defines 18 different types of PHI, ranging from patient names and ID numbers to addresses and phone numbers. The task of removing PHI from a patient note is referred to as \textit{de-identification}.
The essence of de-identification is recognizing PHI in patient notes, which is a form of named-entity recognition (NER).

Existing de-identification systems are often rule-based approaches or feature-based machine learning approaches. However, these techniques require additional lead time for developing and fine-tuning the rules or features specific to each new dataset.
Meanwhile, recent work using ANNs have yielded state-of-the-art performances without using any manual features~\cite{dernoncourt2016identification}. Compared to the previous systems, ANNs have a competitive advantage that the model can be fine-tuned on a new dataset without the overhead of manual feature development, as long as some labels for the dataset are available. 

However, it may still be inefficient to mass deploy ANN-based de-identification system in practical settings, since creating annotations for patient notes is especially difficult. This is due to the fact that only a restricted set of individuals is authorized to access original patient notes; the annotation task cannot be crowd-sourced, making it slow and expensive to obtain a large
annotated corpus.
Medical professionals are therefore wary to explore patient notes because of this de-identification barrier, which considerably hampers medical research.

In this paper, we analyze to what extent transfer learning
may improve de-identification performances on datasets with a limited number of labels.
By training an ANN model on a large dataset (MIMIC) and transferring it to smaller datasets (i2b2 2014 and i2b2 2016), we demonstrate that transfer learning allows to outperform the state-of-the-art results.

\section{Related Work}

Transfer learning has been studied for a long time.
There is no standard definition of transfer learning in the literature~\cite{li2012literature}. We follow the definition from~\cite{pan2010survey}: transfer learning aims at performing a task on a target dataset using some knowledge learned from a source dataset. The idea has been applied to many fields such as speech recognition
~\cite{wang2015transfer} and finance~\cite{stamate2015transfer}.

The successes of ANNs for many applications over the last few years have 
escalated the interest
in studying transfer learning for ANNs. In particular, much work has been done for computer vision~\cite{yosinski2014transferable,oquab2014learning,zeiler2014visualizing}. In these studies, 
some of the parameters learned on the source dataset are used to initialize the corresponding parameters of the ANNs for the target dataset.

Fewer studies have been performed on transfer learning for ANN-based models in the field of natural language processing. For example, Mou et al.~\shortcite{mou2016transferable} focused on transfer learning with convolutional neural networks for sentence classification. To the best of our knowledge, no study has analyzed transfer learning for ANN-based models in the context of NER.

\section{Model}

The model we use for transfer learning experiments is based on a type of recurrent neural networks called long short-term memory (LSTM)~\cite{hochreiter1997long}, and utilizes both token embeddings and character embeddings. 
It comprises six major components:

\begin{enumerate}[itemsep=3pt,topsep=3pt,leftmargin=*]
	\item \textbf{Token embedding layer}  maps each token to a token embedding.
	\item \textbf{Character embedding layer}  maps each character to a character embedding.
	\item \textbf{Character LSTM layer} takes as input character embeddings and outputs a single vector that summarizes the information from the sequence of characters in the corresponding token.
	\item \textbf{Token LSTM layer} takes as input a sequence of token vectors, which are formed by concatenating the outputs of the token embedding layer and the character LSTM layer, and outputs a sequence of vectors.
	\item \textbf{Fully connected layer} takes the output of the token LSTM layer as input, and outputs vectors containing the scores of each label for the corresponding tokens. 
	\item \textbf{Sequence optimization layer} takes the sequence of vectors from the output of the fully connected layer and outputs the most likely sequence of predicted labels, by optimizing the sum of unigram label scores as well as bigram label transition scores.
\end{enumerate}

\begin{figure}[!hb]
  \centering
   \vspace{-0.1cm}
  \includegraphics[width=0.45\textwidth]{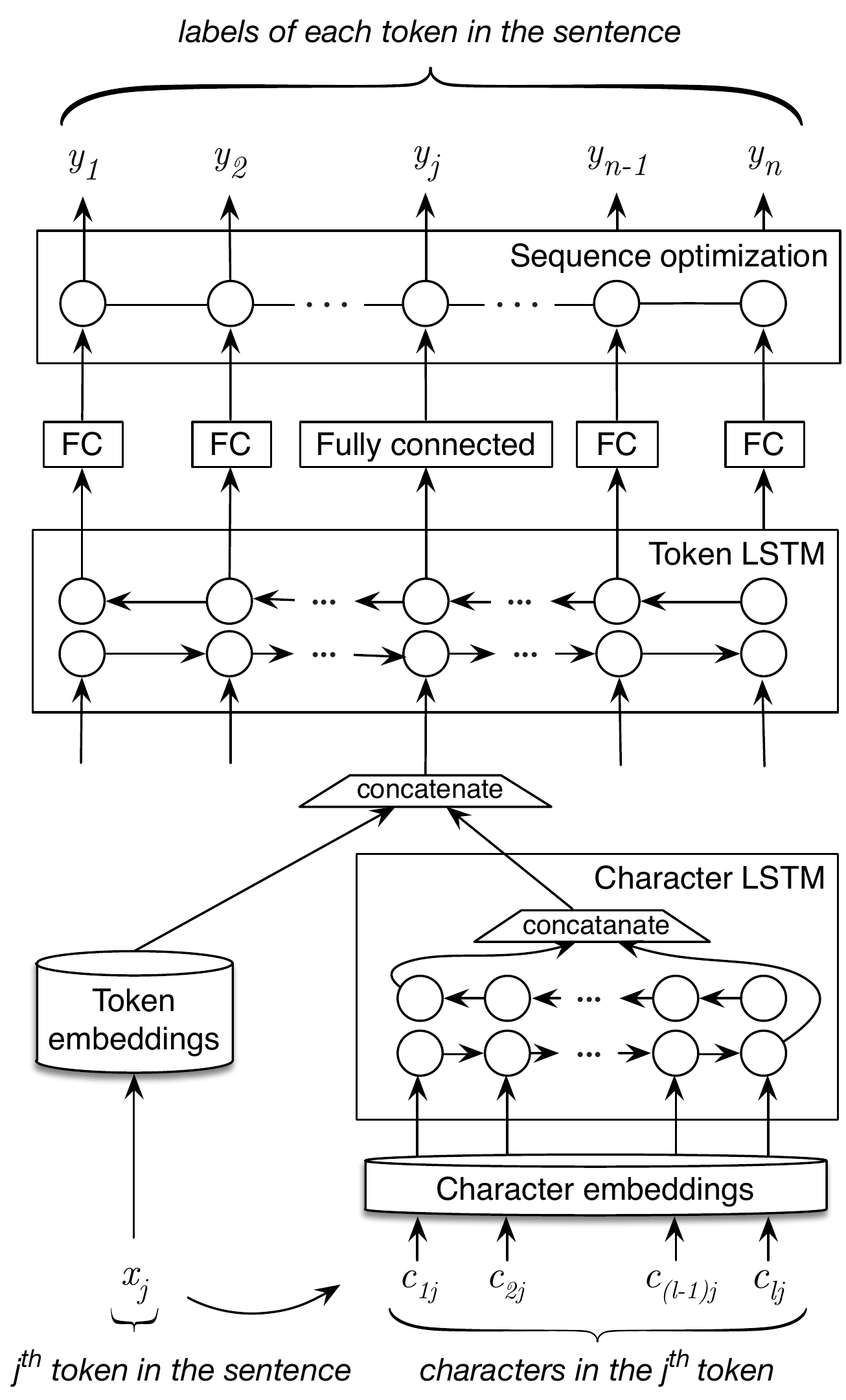}
  \caption{ANN model for NER.
  For transfer learning experiments, we train the parameters of the model on a source dataset, and transfer all or some of the parameters to initialize the model for training on a target dataset.  }

   \vspace{-0.15cm}
  \label{fig:model}
\end{figure}

\noindent Figure~\ref{fig:model} shows how these six components are interconnected to form the model. All layers are learned jointly using stochastic gradient descent. For regularization, dropout is applied before the token LSTM layer, and early stopping is used on the development set with a patience of 10 epochs.

\section{Experiments}

\subsection{Datasets}

We use three de-identification datasets for the transfer learning experiments: MIMIC, i2b2 2014, and i2b2  2016.  
The MIMIC de-identification dataset was introduced in~\cite{dernoncourt2016identification}, and is a subset of the MIMIC-III dataset~\cite{mimic3,goldberger2000physiobank,saeed2011multiparameter}.
The i2b2 2014 and 2016 datasets were released as part of the 2014 i2b2/UTHealth shared task Track~1~\cite{stubbs2015automated} and the 2016 i2b2 CEGS N-GRID shared task, respectively. 
Table~\ref{tab:datasets} presents the datasets' sizes.

\begin{table} [h]
\footnotesize
\centering
\setlength\tabcolsep{2.0pt} % default value: 6pt
\setlength{\extrarowheight}{3pt}
\setlength{\arraycolsep}{5pt}
\begin{tabular}{|l|c|c|c|}

\hline
\textbf{} &  \textbf{MIMIC} & \textbf{i2b2 2014} 	& \textbf{i2b2 2016} 	 \\

\hline
\text{Vocabulary size}		&	69,525& 46,803&61,503\\ 
\text{Number of notes}	& 	1,635&	1,304&1,000\\
\text{Number of tokens}	& 2,945,228	&	984,723 &2,689,196\\  
\text{Number of PHI instances}		& 60,725&	28,867& 41,142\\  
\text{Number of PHI tokens}	& 78,633&41,355	&54,420	\\  
 
\hline
\end{tabular}
\caption{Overview of the MIMIC and i2b2 datasets. PHI stands for protected health information.}
\label{tab:datasets}
\end{table}

\vspace{-0.5cm}

\subsection{Transfer learning}

The goal of transfer learning is to leverage the information present in a source dataset to improve the performance of an algorithm on a target dataset. In our setting,  we apply transfer  learning  by training  the parameters  of the ANN model on the source dataset (MIMIC), and using the same ANN to retrain on the target dataset (i2b2 2014 or 2016) for fine-tuning. We use MIMIC as the source dataset since it is the dataset with the most labels.  We perform two sets of experiments to gain insights on how effective transfer learning is and which parameters of the ANN are the most important to transfer.\footnote{Our code is an extension of the NER library {NeuroNER}~\cite{2017neuroner}, which we committed to NeuroNER's repository \url{https://github.com/Franck-Dernoncourt/NeuroNER}}

\paragraph{Experiment 1} Quantifying the impact of transfer learning for various train set sizes of the target dataset. The primary purpose of this experiment is to assess to what extent transfer learning improves the performances on the target dataset.  We experiment with different train set sizes to understand how many labels are needed for the target dataset to achieve reasonable performances with and without transfer learning.

\paragraph{Experiment 2} Analyzing the importance of each parameter of the ANN
in the transfer learning. Instead of transferring all the parameters, we experiment with transferring different combinations of parameters. The goal is to understand which components of the ANN are the most important to transfer. The lowest layers of the ANN tend to represent task-independent features, whereas the topmost layers are more task-specific. As a result, we try transferring the parameters starting from the bottommost layer up to the topmost layer, adding one layer at a time.

\section{Results}

\begin{figure*}[ht]
\vspace{-0.3cm}
\centering
\hspace{-0.1cm}
\captionsetup[subfigure]{margin={1cm,0.4cm}}
  \subfloat[\vspace{-0.0cm}][i2b2 2014]{\includegraphics[height=4.9cm]{{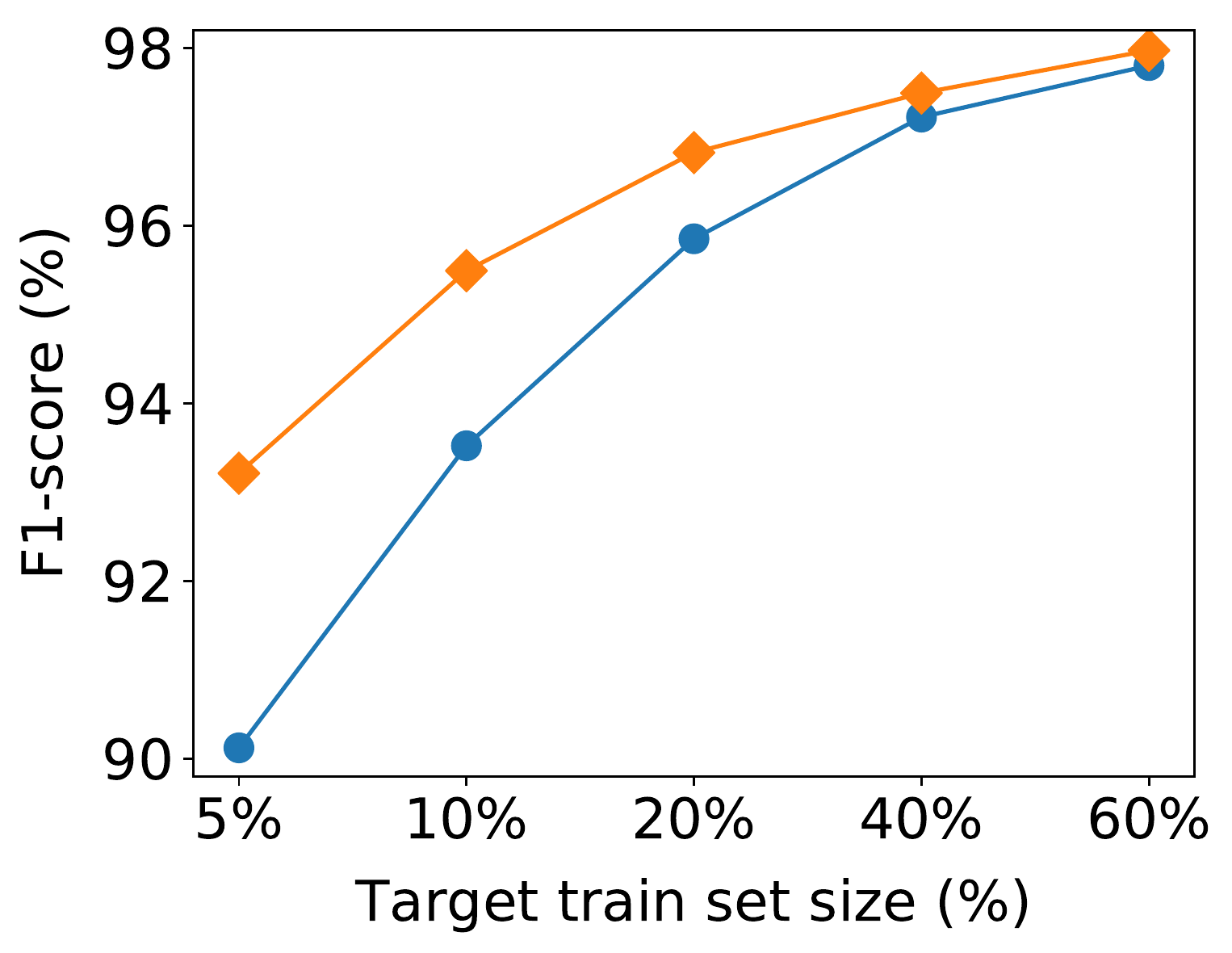}} \label{fig:tf-vs-baseline_i2b2-2014}} \hspace{0.4cm}
\captionsetup[subfigure]{margin={0cm,2cm}}
    \subfloat[\vspace{-0.0cm}][i2b2 2016]{\includegraphics[height=4.9cm]{{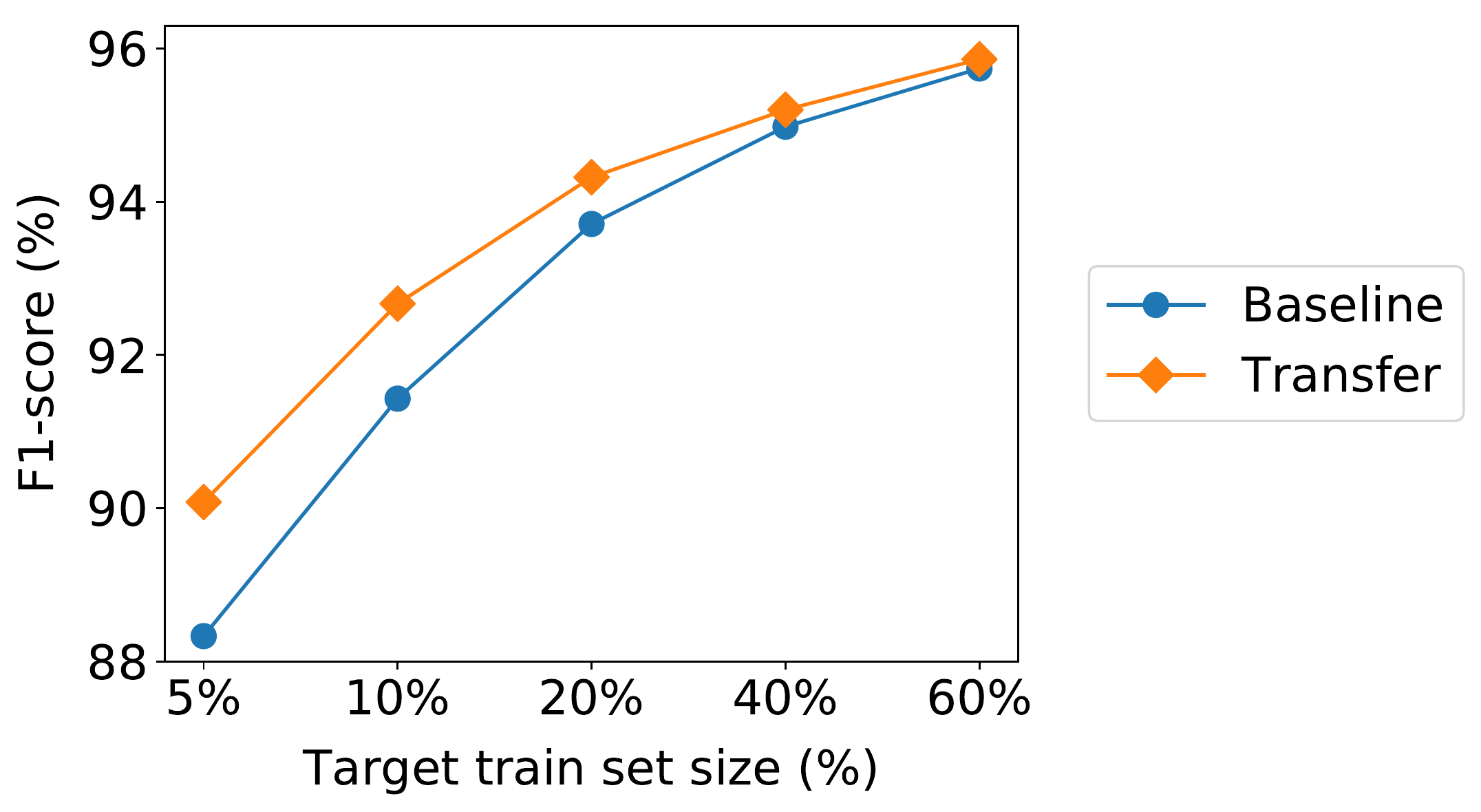}}  \label{fig:tf-vs-baseline_i2b2-2016}} \hspace{0.4cm}%

  \vspace{-0.0cm}
  \caption{Impact of transfer learning on the F1-scores. Baseline corresponds to training the ANN model only with the target dataset, and transfer learning corresponds to training on the source dataset followed by training on the target dataset. The target train set size is the percentage of train set in the whole dataset, and 60\% corresponds to the full official train set.} \label{fig:tf-vs-baseline}
  \vspace{-0.0cm}
\end{figure*}

\paragraph{Experiment 1}
Figure~\ref{fig:tf-vs-baseline} compares the F1-scores of the ANN trained only on the target dataset against the ANN trained on the source dataset followed by the target dataset.
Transfer learning improves the F1-scores over training only with the target  dataset, though the improvement diminishes as the number of training samples used for the target dataset increases. This implies that the representations learned from the source dataset are efficiently transferred and exploited for the target dataset. 

Therefore, when transfer learning is adopted, fewer annotations are needed to achieve the same level of performance as when the source dataset is unused.
For example, on the i2b2 2014 dataset, performing transfer learning and using 16\% of the i2b2 train set leads to similar performance as not using transfer learning and using 34\% of the i2b2 train set. 
Transfer learning thus allows to cut by half the number of labels needed on the target dataset in this case.

For both the i2b2 2014 and 2016 datasets, the performance gains from transfer learning are greater when the train set size of the target dataset is small. 
The largest improvement can be observed for i2b2 2014 when using 5\% of the dataset as the train set (consisting of around 2k PHI tokens out of 50k tokens), where transfer learning increases the F1-score by around 3.1 percent point, from 90.12 to 93.21. Even when all of the train set is used, the F1-score improves when using transfer learning, 
albeit by just 0.17 percent point, from 97.80 to 97.97.

\begin{figure*}[ht]
\vspace{-0.2cm}
\centering
\hspace{-0.3cm}
\captionsetup[subfigure]{margin={1cm,0.2cm}}
  \subfloat[\vspace{-0.0cm}][i2b2 2014]{\includegraphics[height=6.1cm]{{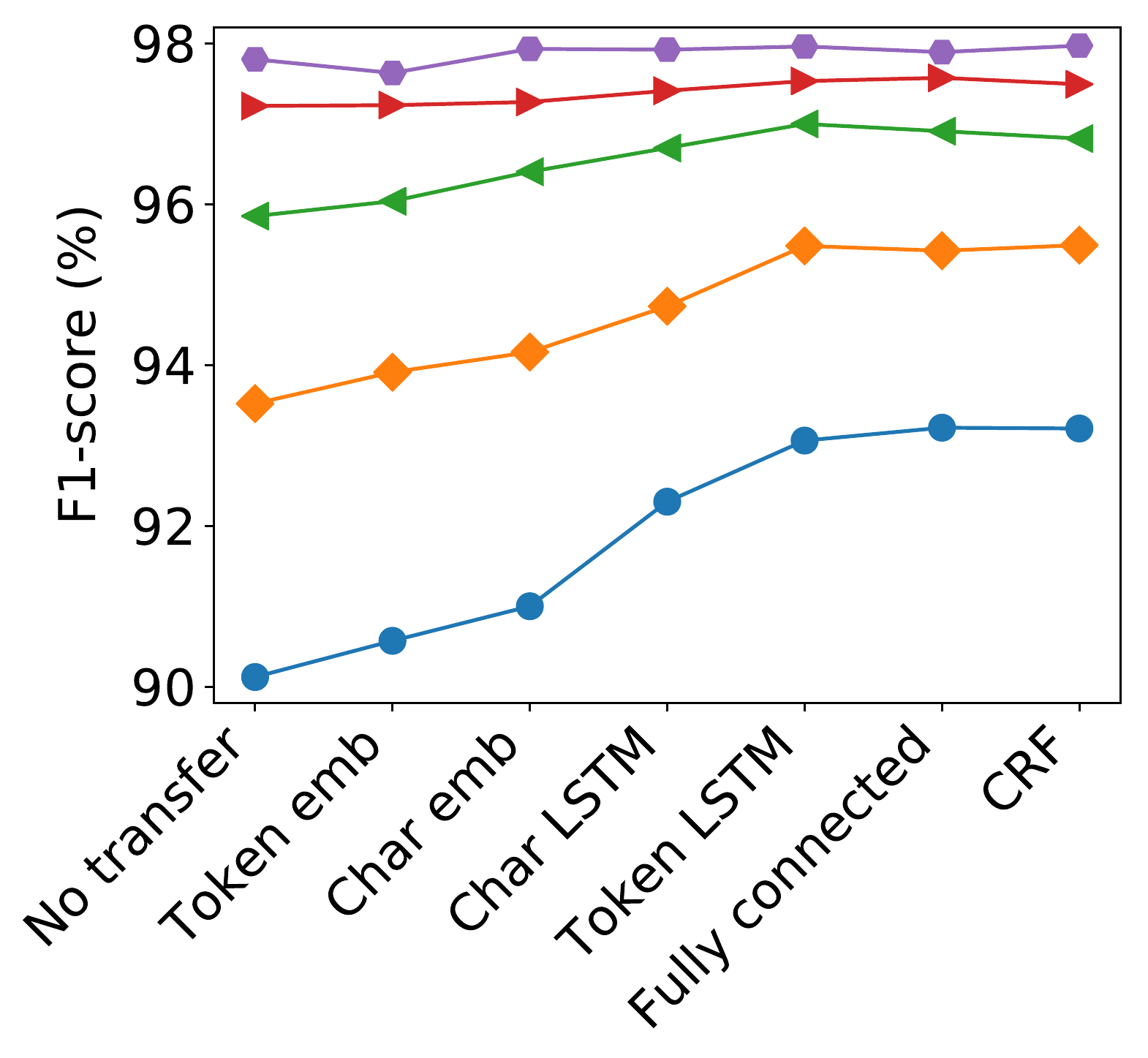}}\label{fig:tf-per-layer_i2b2-2014}} \hspace{0.6cm}
\captionsetup[subfigure]{margin={0.5cm,2cm}}
    \subfloat[\vspace{-0.0cm}][i2b2 2016]{\includegraphics[height=6.1cm]{{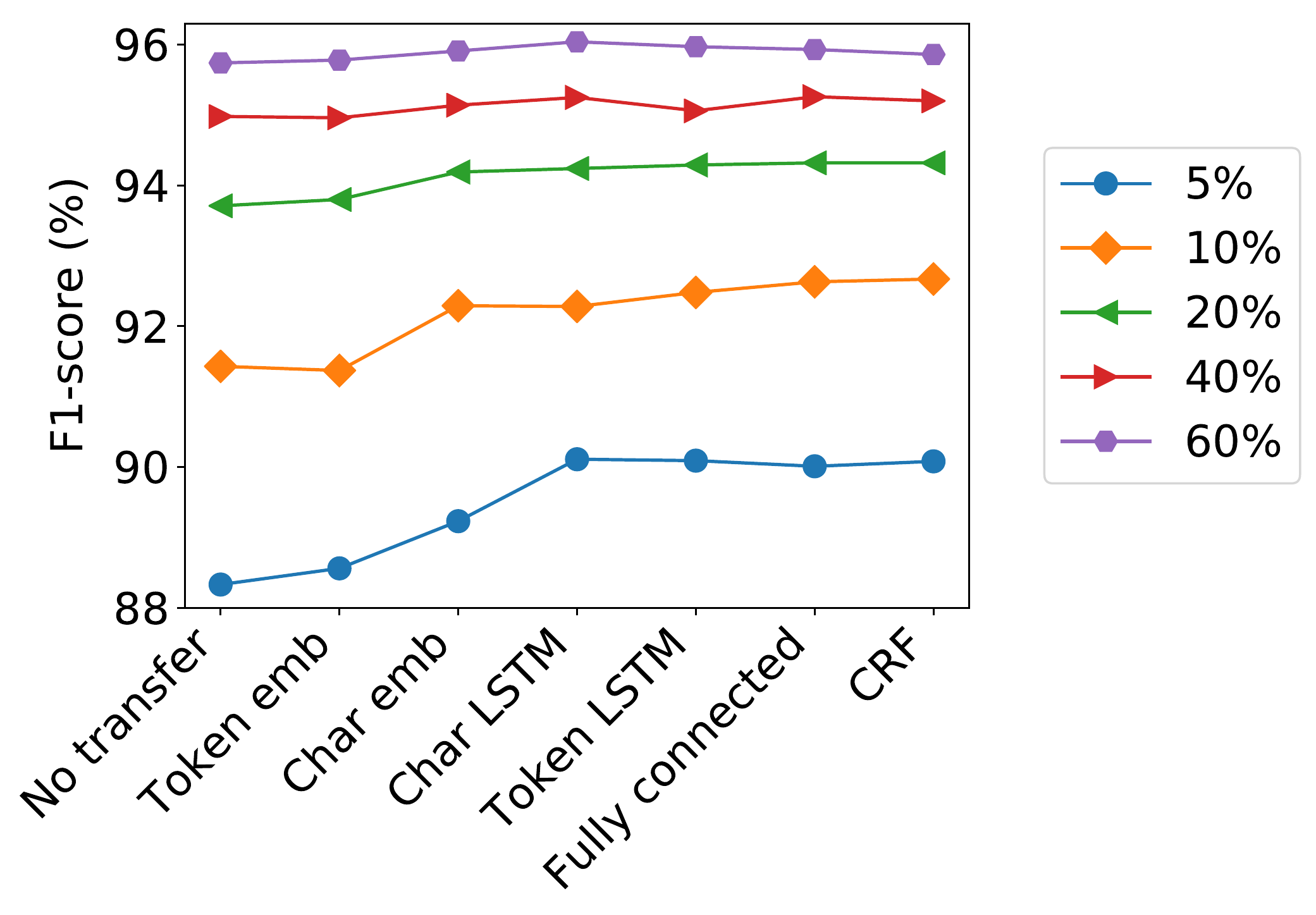}} \label{fig:tf-per-layer_i2b2-2016}} \hspace{0.4cm}%
    \hspace{0.2cm}

  \vspace{-0.2cm}
  \caption{Impact of transferring the parameters up to each layer of the ANN model using various train set sizes on the target dataset: 5\%, 10\%, 20\%, 40\%, and 60\% (official train set).} 
\label{fig:tf-per-layer}
  \vspace{-0.1cm}
\end{figure*}

\paragraph{Experiment 2}
Figure~\ref{fig:tf-per-layer} shows the importance of each layer of the ANN in transfer learning.
We observe that transferring a few lower layers is almost as efficient as transferring all layers. For i2b2 2014, transferring up to the token LSTM shows great improvements for each layer, but there is less improvement for each added layer beyond that. For i2b2 2016, larger improvements can be observed up to the character LSTM and less so beyond that layer.

The parameters in the lower layers therefore seems to contain most information that are relevant to the de-identification task in general, which supports the common hypothesis that higher layers of ANN architectures contain the parameters that are more specific to the task as well as the dataset used for training.

Despite the observation that transferring a few lower layers may be sufficient for efficient transfer learning, it is
interesting to see that
adding the topmost layers to the transfer learning does not hurt the performance. When retraining the model on the target dataset, the ANN is able to adapt
to the target dataset quite well despite some the higher layers being initialized to parameters that are likely to be more specific to the source dataset.

\section{Conclusion}

In this work, we have studied transfer learning with ANNs 
for NER, specifically patient note de-identification, 
by transferring ANN parameters trained on a large labeled dataset to another dataset with limited human annotations. We demonstrated that transfer learning improves the performance over the state-of-the-art results on two 
datasets.
Transfer learning may be especially beneficial for a target dataset with small number of labels.

\bibliography{xample-arxiv}

\newcommand{\noopsort}[1]{} \newcommand{\printfirst}[2]{#1}
  \newcommand{\singleletter}[1]{#1} \newcommand{\switchargs}[2]{#2#1}
\begin{thebibliography}{}
\expandafter\ifx\csname natexlab\endcsname\relax\def\natexlab#1{#1}\fi

\bibitem[{Ananthakrishnan et~al.(2013)Ananthakrishnan, Cai, Savova, Cheng,
  Chen, Perez, Gainer, Murphy, Szolovits, Xia
  et~al.}]{ananthakrishnan2013improving}
Ashwin~N Ananthakrishnan, Tianxi Cai, Guergana Savova, Su-Chun Cheng, Pei Chen,
  Raul~Guzman Perez, Vivian~S Gainer, Shawn~N Murphy, Peter Szolovits, Zongqi
  Xia, et~al. 2013.
\newblock Improving case definition of {Crohn's} disease and ulcerative colitis
  in electronic medical records using natural language processing: a novel
  informatics approach.
\newblock {\em Inflammatory bowel diseases\/} 19(7):1411.

\bibitem[{Dernoncourt et~al.(2017)Dernoncourt, Lee, and
  Szolovits}]{2017neuroner}
Franck Dernoncourt, Ji~Young Lee, and Peter Szolovits. 2017.
\newblock {NeuroNER}: an easy-to-use program for named-entity recognition based
  on neural networks.
\newblock {\em arXiv:1705.05487\/} .

\bibitem[{Dernoncourt et~al.(2016)Dernoncourt, Lee, Uzuner, and
  Szolovits}]{dernoncourt2016identification}
Franck Dernoncourt, Ji~Young Lee, Ozlem Uzuner, and Peter Szolovits. 2016.
\newblock De-identification of patient notes with recurrent neural networks.
\newblock {\em Journal of the American Medical Informatics Association\/} page
  ocw156.

\bibitem[{Goldberger et~al.(2000)Goldberger, Amaral, Glass, Hausdorff, Ivanov,
  Mark, Mietus, Moody, Peng, and Stanley}]{goldberger2000physiobank}
Ary~L Goldberger, Luis~AN Amaral, Leon Glass, Jeffrey~M Hausdorff, Plamen~Ch
  Ivanov, Roger~G Mark, Joseph~E Mietus, George~B Moody, Chung-Kang Peng, and
  H~Eugene Stanley. 2000.
\newblock Physiobank, physiotoolkit, and physionet components of a new research
  resource for complex physiologic signals.
\newblock {\em Circulation\/} 101(23):e215--e220.

\bibitem[{Halpern et~al.(2016)Halpern, Horng, Choi, and
  Sontag}]{halpern2016electronic}
Yoni Halpern, Steven Horng, Youngduck Choi, and David Sontag. 2016.
\newblock Electronic medical record phenotyping using the anchor and learn
  framework.
\newblock {\em Journal of the American Medical Informatics Association\/} page
  ocw011.

\bibitem[{Hochreiter and Schmidhuber(1997)}]{hochreiter1997long}
Sepp Hochreiter and J{\"u}rgen Schmidhuber. 1997.
\newblock Long short-term memory.
\newblock {\em Neural computation\/} 9(8):1735--1780.

\bibitem[{Johnson et~al.(2016)Johnson, Pollard, Shen, wei Lehman, Feng,
  Ghassemi, Moody, Szolovits, Celi, and Mark}]{mimic3}
Alistair E.~W. Johnson, Tom~J. Pollard, Lu~Shen, Li~wei Lehman, Mengling Feng,
  Mohammad Ghassemi, Benjamin Moody, Peter Szolovits, Leo~Anthony Celi, and
  Roger~G. Mark. 2016.
\newblock {MIMIC-III}, a freely accessible critical care database.
\newblock {\em Scientific Data\/} .

\bibitem[{Li(2012)}]{li2012literature}
Qi~Li. 2012.
\newblock Literature survey: domain adaptation algorithms for natural language
  processing.
\newblock {\em Department of Computer Science The Graduate Center, The City
  University of New York\/} pages 8--10.

\bibitem[{Liao et~al.(2015)Liao, Cai, Savova, Murphy, Karlson, Ananthakrishnan,
  Gainer, Shaw, Xia, Szolovits et~al.}]{liao2015development}
Katherine~P Liao, Tianxi Cai, Guergana~K Savova, Shawn~N Murphy, Elizabeth~W
  Karlson, Ashwin~N Ananthakrishnan, Vivian~S Gainer, Stanley~Y Shaw, Zongqi
  Xia, Peter Szolovits, et~al. 2015.
\newblock Development of phenotype algorithms using electronic medical records
  and incorporating natural language processing.
\newblock {\em bmj\/} 350:h1885.

\bibitem[{Mou et~al.(2016)Mou, Meng, Yan, Li, Xu, Zhang, and
  Jin}]{mou2016transferable}
Lili Mou, Zhao Meng, Rui Yan, Ge~Li, Yan Xu, Lu~Zhang, and Zhi Jin. 2016.
\newblock How transferable are neural networks in {NLP} applications?
\newblock {\em arXiv preprint arXiv:1603.06111\/} .

\bibitem[{Murdoch and Detsky(2013)}]{murdoch2013inevitable}
Travis~B Murdoch and Allan~S Detsky. 2013.
\newblock The inevitable application of big data to health care.
\newblock {\em Jama\/} 309(13):1351--1352.

\bibitem[{Office~for Civil~Rights(2002)}]{office2002standards}
HHS Office~for Civil~Rights. 2002.
\newblock Standards for privacy of individually identifiable health
  information. final rule.
\newblock {\em Federal Register\/} 67(157):53181.

\bibitem[{Oquab et~al.(2014)Oquab, Bottou, Laptev, and
  Sivic}]{oquab2014learning}
Maxime Oquab, Leon Bottou, Ivan Laptev, and Josef Sivic. 2014.
\newblock Learning and transferring mid-level image representations using
  convolutional neural networks.
\newblock In {\em Proceedings of the IEEE conference on computer vision and
  pattern recognition\/}. pages 1717--1724.

\bibitem[{Pan and Yang(2010)}]{pan2010survey}
Sinno~Jialin Pan and Qiang Yang. 2010.
\newblock A survey on transfer learning.
\newblock {\em IEEE Transactions on knowledge and data engineering\/}
  22(10):1345--1359.

\bibitem[{Pivovarov and Elhadad(2015)}]{pivovarov2015automated}
Rimma Pivovarov and No{\'e}mie Elhadad. 2015.
\newblock Automated methods for the summarization of electronic health records.
\newblock {\em Journal of the American Medical Informatics Association\/}
  22(5):938--947.

\bibitem[{Saeed et~al.(2011)Saeed, Villarroel, Reisner, Clifford, Lehman,
  Moody, Heldt, Kyaw, Moody, and Mark}]{saeed2011multiparameter}
Mohammed Saeed, Mauricio Villarroel, Andrew~T Reisner, Gari Clifford, Li-Wei
  Lehman, George Moody, Thomas Heldt, Tin~H Kyaw, Benjamin Moody, and Roger~G
  Mark. 2011.
\newblock Multiparameter intelligent monitoring in intensive care {II}
  ({MIMIC-II}): a public-access intensive care unit database.
\newblock {\em Critical care medicine\/} 39(5):952.

\bibitem[{Stamate et~al.(2015)Stamate, Magoulas, and
  Thomas}]{stamate2015transfer}
Cosmin Stamate, George~D Magoulas, and Michael~SC Thomas. 2015.
\newblock Transfer learning approach for financial applications.
\newblock {\em arXiv preprint arXiv:1509.02807\/} .

\bibitem[{Stubbs et~al.(2015)Stubbs, Kotfila, and Uzuner}]{stubbs2015automated}
Amber Stubbs, Christopher Kotfila, and {\"O}zlem Uzuner. 2015.
\newblock Automated systems for the de-identification of longitudinal clinical
  narratives: Overview of 2014 i2b2/{UTHealth} shared task track 1.
\newblock {\em Journal of biomedical informatics\/} 58:S11--S19.

\bibitem[{Wang and Zheng(2015)}]{wang2015transfer}
Dong Wang and Thomas~Fang Zheng. 2015.
\newblock Transfer learning for speech and language processing.
\newblock In {\em Signal and Information Processing Association Annual Summit
  and Conference (APSIPA), 2015 Asia-Pacific\/}. IEEE, pages 1225--1237.

\bibitem[{Yosinski et~al.(2014)Yosinski, Clune, Bengio, and
  Lipson}]{yosinski2014transferable}
Jason Yosinski, Jeff Clune, Yoshua Bengio, and Hod Lipson. 2014.
\newblock How transferable are features in deep neural networks?
\newblock In {\em Advances in neural information processing systems\/}. pages
  3320--3328.

\bibitem[{Zeiler and Fergus(2014)}]{zeiler2014visualizing}
Matthew~D Zeiler and Rob Fergus. 2014.
\newblock Visualizing and understanding convolutional networks.
\newblock In {\em European conference on computer vision\/}. Springer, pages
  818--833.

\end{thebibliography}
\bibliographystyle{emnlp_natbib}

\end{document}